\documentclass[10pt, journal]{IEEEtran}
\usepackage{amsmath,amsfonts}
\usepackage{algorithm}
\usepackage{graphicx}
\usepackage{algorithm}
\usepackage{array}
\usepackage[caption=false,font=normalsize,labelfont=sf,textfont=sf]{subfig}
\usepackage{textcomp}
\usepackage{xcolor}
\usepackage{algpseudocode}
\usepackage{amsfonts}
\usepackage{amsthm}
\usepackage{stfloats}
\usepackage{subcaption}
\usepackage{booktabs}
\usepackage{tabularx}
\def\BibTeX{{\rm B\kern-.05em{\sc i\kern-.025em b}\kern-.08em
    T\kern-.1667em\lower.7ex\hbox{E}\kern-.125emX}}

\begin{document}

\title{Adaptive Federated LoRA in Heterogeneous Wireless Networks with Independent Sampling}
\author{Yanzhao Hou,~\IEEEmembership{Member,~IEEE}, Jiaxiang Geng,~\IEEEmembership{Student Member, ~IEEE}, Boyu Li,~\IEEEmembership{Student Member,~IEEE}, \\Xiaofeng Tao,~\IEEEmembership{Senior Member,~IEEE}, Juncheng Wang,~\IEEEmembership{Member,~IEEE}, Xiaodong Xu,~\IEEEmembership{Senior Member,~IEEE}, \\and Bing Luo,~\IEEEmembership{Senior Member,~IEEE}
    \thanks{
    Y. Hou, X. Xu and X. Tao are with the National Engineering Research Center for Mobile Network Technologies,  Beijing University of Posts and Telecommunications, Beijing, 100876, China (e-mail: \{houyanzhao, xuxiaodong, taoxf\}@bupt.edu.cn).
    
    J. Geng and B. Li are with the National Engineering Research Center for Mobile Network Technologies, Beijing University of Posts and Telecommunications, Beijing, 100876, China (e-mail: \{lelegjx, liboyu\}@bupt.edu.cn).
    
    J. Wang is with the Hong Kong Baptist University, Hong Kong, China (e-mail: jcwang@comp.hkbu.edu.hk).
    
    B. Luo is affiliated with Peng Cheng Laboratory. This work was conducted while he was a Visiting Scholar at Peng Cheng Laboratory (e-mail: luobing1008@gmail.com).
    
    Corresponding author: Yanzhao Hou and Bing Luo}
}

\markboth{Journal of \LaTeX\ Class Files,~Vol.~14, No.~8, August~2021}%
{Shell \MakeLowercase{\textit{et al.}}: A Sample Article Using IEEEtran.cls for IEEE Journals}

\IEEEpubid{}

\maketitle

\begin{abstract}
Federated LoRA has emerged as a promising technique for efficiently fine-tuning large language models (LLMs) on distributed devices by reducing the number of trainable parameters. However, existing approaches often inadequately overlook the theoretical and practical implications of system and data heterogeneity, thereby failing to optimize the overall training efficiency, particularly in terms of wall-clock time. In this paper, we propose an adaptive federated LoRA strategy with independent client sampling to minimize the convergence wall-clock time of federated fine-tuning under both computation and communication heterogeneity. We first derive a new convergence bound for federated LoRA with arbitrary and independent client sampling, notably without requiring the stringent bounded gradient assumption. Then, we introduce an adaptive bandwidth allocation scheme that accounts for heterogeneous client resources and system bandwidth constraints. Based on the derived theory, we formulate and solve a non-convex optimization problem to jointly determine the LoRA sketching ratios and sampling probabilities, aiming to minimize wall-clock convergence time. An efficient and low-complexity algorithm is developed to approximate the solution. Finally, extensive experiments demonstrate that our approach significantly reduces wall-clock training time compared to state-of-the-art methods across various models and datasets.

\end{abstract}

\begin{IEEEkeywords}
LLM Fine-tuning, Federated Learning, LoRA, Client Sampling, Wireless Network.
\end{IEEEkeywords}

\section{Introduction}
\IEEEPARstart{L}{arge} Language Models (LLMs), such as GPT \cite{radford2018improving, radford2019language} and LLaMA \cite{touvron2023llamaopenefficientfoundation}, as well as more recent architectures \cite{achiam2023gpt, team2024gemini, liu2024deepseek} have revolutionized natural language processing by leveraging pre-training on massive and diverse corpora to learn broad and transferable representations. While these foundation models demonstrate strong zero-shot and few-shot capabilities across various downstream tasks, adapting them to specific user needs or domain-specific applications often requires fine-tuning, which typically depends on large amounts of data \cite{sun2019fine}. With the rapid development of 5G, the Internet of Things (IoT), and social networking applications, massive volumes of data are now being generated at the wireless network edge \cite{b1}. Traditional centralized fine-tuning of LLMs requires transmitting this edge data to a central server, which raises significant privacy concerns \cite{b2}. Furthermore, transferring such large-scale data consumes substantial bandwidth, making centralized approaches both inefficient and impractical in many real-world scenarios \cite{10944288}.

As an appealing distributed machine learning (DML) paradigm, Federated Learning (FL) enables multiple local clients to collaboratively fine-tune a global model without sharing their raw data \cite{mcmahan2017communication}. By keeping data decentralized, FL effectively preserves user privacy and reduces communication overhead. This approach has gained widespread adoption across various domains, including computer vision, natural language processing, e-healthcare, and smart home applications \cite{b3,b5,fedcampus}. However, enabling on-device fine-tuning of LLMs in federated learning remains challenging \cite{10944288}. LLMs typically contain billions of parameters, leading to substantial computational, communication, and memory overhead during training. Full-parameter fine-tuning is often beyond the capabilities of edge devices with limited resources~\cite{10447454}.

Fortunately, parameter-efficient fine-tuning methods have been proposed to reduce the computational and storage costs of fine-tuning \cite{powerofscale, LoRA, li-liang-2021-prefix}. Among the methods, Low-Rank Adaptation (LoRA) is a widely used approach due to its flexibility that significantly reduces the number of trainable parameters by injecting small trainable matrices into specific layers of the model, which allows the model to be efficiently adapted to new tasks without modifying the original pre-trained weights \cite{LoRA}. To enable distributed on-device fine-tuning of LLMs, recent studies have integrated LoRA with the FL algorithm, forming a framework commonly referred to as Federated LoRA \cite{openfedllm}.

Federated LoRA exhibits two distinctive characteristics. First, the data distribution across clients is often highly non-independent and identically distributed (non-i.i.d.) and unbalanced, a phenomenon known as data heterogeneity, which can significantly hinder the convergence performance of FL algorithms \cite{b20}. Second, local clients typically possess varying levels of communication and computational capabilities, referred to as system heterogeneity, which leads to the straggler problem and slows down the overall training process \cite{b21}. This challenge becomes even more pronounced in wireless edge networks, where limited and shared communication bandwidth further exacerbates the impact of system heterogeneity \cite{b25}.

\subsection{Related Works}

\subsubsection{FL methods for improving fine-tuning efficiency}
\
\newline
\indent To address the network heterogeneity and improve communication efficiency, FL algorithms such as the standard FedAvg algorithm typically sample a \textit{subset} of clients and perform multiple local model updates \cite{b8}. Recent studies have provided theoretical convergence analysis for FL with \emph{client sampling}\footnote{Client sampling here means that all clients will participate in FL throughout the training rounds, with certain probability of participation in each round (not to casually sample a fixed subset of clients in each round), which guarantees the model convergence and unbiasedness.} \cite{b12,b22,b23}. However, these works have primarily focused on the data heterogeneity but neglecting the system heterogeneity, and therefore often exhibit slow convergence in terms of the wall-clock time. The reason is that these works may select a straggler under poor wireless channel condition, resulting in a long per-round time \cite{9488679}. 

In addressing both data and system heterogeneity, recent works \cite{b15,b,b26} have designed client sampling strategies in wireless networks to minimize FL convergence time. However, the convergence analysis in \cite{b15} is only valid for convex loss function. Furthermore, their sampling probabilities among clients are \emph{dependent}, and a \emph{fixed} number of participating clients are selected in each round. In practice, some FL clients may experience unexpected disconnections due to the fluctuation of wireless networks, while others may be unable to participate in FL training due to their higher-priority tasks, which make the participation of each client in FL training \emph{independent}. Although \cite{b} and \cite{b26} studied independent client sampling, they only considered the communication heterogeneity while neglecting the computation heterogeneity. As shown in Fig. 1, when FL clients perform multiple local iterations, computation heterogeneity also has a significant impact on the training round time, and should be jointly designed and optimized with the communication heterogeneity. Moreover, their convergence results rely on a uniform bounded gradient assumption, which may result in a loose bound, since the gradient norm usually decays over the training rounds. 

\vspace{2mm}
\subsubsection{Federated LoRA methods}
\
\newline
\indent Federated LoRA is emerging as a promising solution for collaborative fine-tuning of LLMs across distributed devices \cite{chen2024integrationlargelanguagemodels}. Recent efforts have extended this framework in various directions. For instance, differential privacy has been incorporated  to enhance security \cite{sun2024improving}, and communication compression has been applied  to reduce transmission overhead \cite{kuo2024federatedlorasparsecommunication}. Other methods have addressed the challenge of system heterogeneity by proposing heterogeneous LoRA mechanisms in \cite{bai2024federated,hetlora,byun-lee-2025-towards,flora,koo2024robustefficientfederatedlowrank,fedsketchlora}, allowing devices to use different LoRA ranks.

However, most of these approaches lack theoretical analysis and fail to provide a comprehensive understanding of their impact on overall system efficiency, particularly in terms of wall-clock time. For example, FlexLoRA \cite{bai2024federated} applies truncated singular value decomposition (SVD) to align heterogeneous LoRA modules, but this results in additional computation and memory overhead, potentially limiting scalability. HeteroLoRA \cite{hetlora} uses zero-padding to align update dimensions across devices, but this ad hoc approach introduces redundancy and lacks optimization guarantees. ReplicationLoRA \cite{byun-lee-2025-towards} analyzes the noise error introduced by zero-padding in HeteroLoRA and proposes a replication strategy to mitigate it, but it lacks a convergence analysis. FedStackLoRA \cite{flora} stacks LoRA modules across devices, which increases communication cost linearly with the number of participants and compromises the modularity and flexibility that LoRA was originally designed to support. LoRA-$A^2$ \cite{koo2024robustefficientfederatedlowrank} demonstrates robustness in challenging settings with low ranks and high data heterogeneity, but overlooks the impact of system heterogeneity on overall system efficiency. FSLoRA \cite{fedsketchlora} proposes a sketching mechanism to reduce update size, but it does not address how to optimally choose LoRA sketching ratios under system heterogeneity to minimize wall-clock time.

Crucially, none of these methods systematically consider the interplay between LoRA parameters' sizes, client sampling, data heterogeneity and system heterogeneity to optimize the overall fine-tuning efficiency. Despite their empirical benefits, the absence of theoretical grounding limits their applicability to practical, large-scale, resource-constrained FL systems.

\begin{figure*}
\centering
\includegraphics[width=0.9\textwidth]{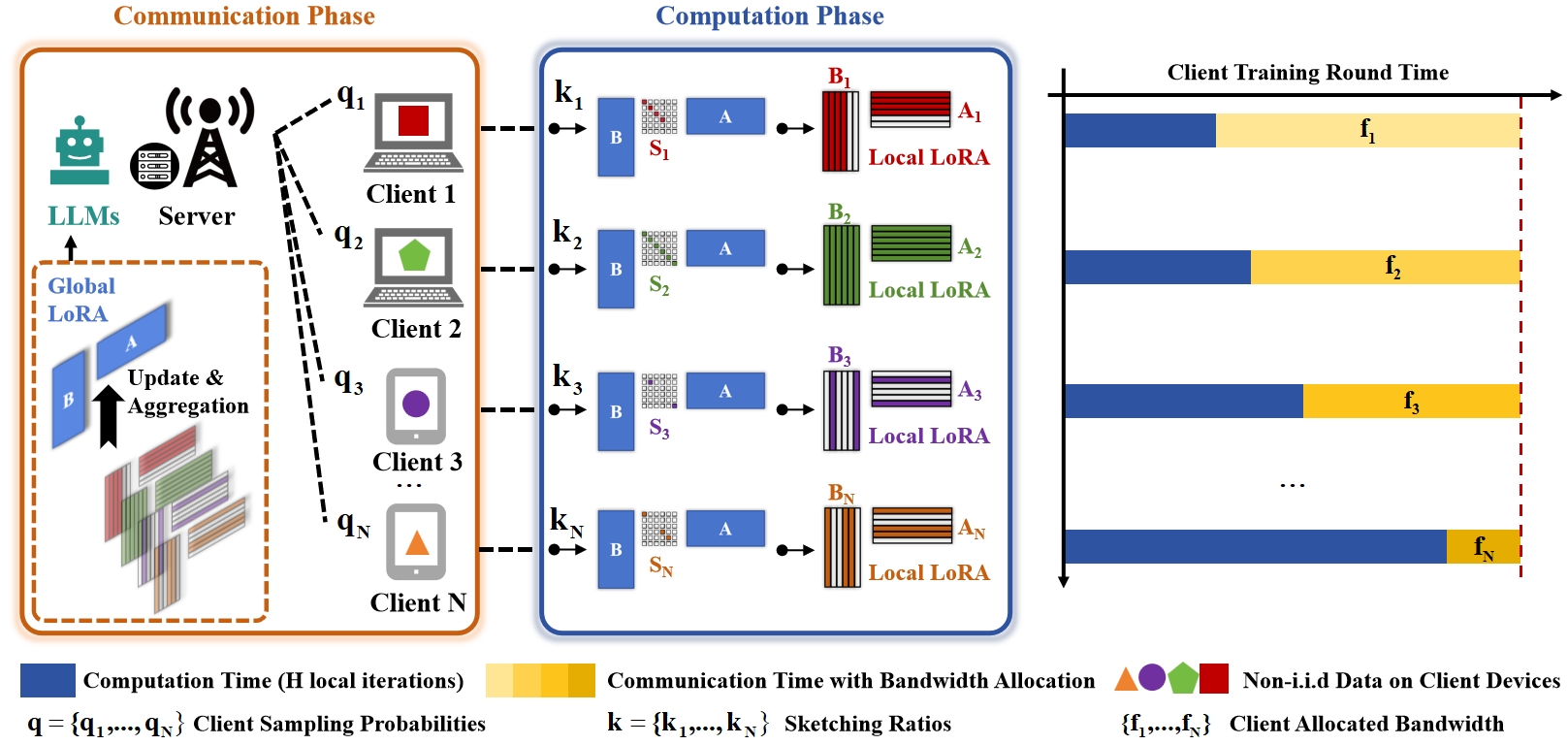}
\caption{A heterogeneous federated LoRA over wireless networks, where, clients with heterogeneous computation and communication capacities, as well as data distribution.}
\label{fig:overview}
\end{figure*}

\subsection{Our Contributions}

Motivated by the above discrepancies, in this work, we propose a new \emph{adaptive federated LoRA strategy with independent client sampling} to minimize the FL convergence wall-clock time, while considering data heterogeneity and system heterogeneity in \textit{both} communication and computation. The main contributions of this paper are as follows:

\begin{itemize}
\item Based on the sketching mechanism, we derive a new convergence bound for federated LoRA with arbitrary sketching ratios and independent client sampling probabilities. Our convergence bound holds for general non-convex loss function without the stringent bounded gradient assumptions.
\item We propose an adaptive bandwidth allocation scheme for Federated LoRA with independent client sampling. Our proposed approach jointly considers the impacts of heterogeneous computation and communication capacities, as well as the limited system bandwidth, to characterize the expected wall-clock time for each training round.
\item Building upon the derived convergence bound and the proposed adaptive bandwidth allocation scheme, we formulate an optimization problem on the LoRA sketching ratios and independent sampling probabilities, to minimize the federated LoRA convergence wall-clock time with both data and system heterogeneity. We develop an efficient algorithm to approximately solve the non-convex optimization problem with low computational complexity.
\item We evaluate the performance of our proposed algorithm under heterogeneous network settings.
Our experimental results under various learning models and datasets demonstrate that the proposed algorithm substantially reduces the federated LoRA wall-clock finetuning time compared with state-of-the-art methods.
\end{itemize}

\section{System Model}
\subsection{Federated LoRA Fine-tuning with sketching mechanism}

LoRA enables efficient fine-tuning by approximating weight updates $\Delta \mathbf{W}$ using a low-rank decomposition $\Delta \mathbf{W} = \mathbf{B}\mathbf{A}$, where $\mathbf{B}$ and $\mathbf{A}$ are small trainable matrices that contain significantly fewer parameters than the original weight matrix \cite{LoRA}.
As illustrated in Fig. 1, we consider a federated LoRA system operating at the wireless edge, which consists of a set of $N$ clients indexed by $\mathcal{N} = \{1, \ldots, N\}$ and a central server. The federated LoRA fine-tuning problem can then be formulated as:
\begin{equation}
\begin{aligned}
\min_{\mathbf{B,A}} F(\mathbf{B},\mathbf{A}) &= \sum_{n=1}^N a_n F_n(\mathbf{B},\mathbf{A}) \\
\text{where} \quad F_n(\mathbf{B},\mathbf{A}) &= \mathbb{E}_{d\sim \mathcal{D}_n} [\mathcal{L}(\mathbf{W}_0 + \mathbf{BA};d)]. 
\end{aligned}
\end{equation}

Here, $\mathbf{W}_0$ denotes the frozen (pre-trained) base model. The LoRA modules are represented by $\mathbf{B} \in \mathbb{R}^{m \times \gamma}$ and $\mathbf{A} \in \mathbb{R}^{\gamma \times n}$, which introduce a low-rank update $\Delta \mathbf{W} = \mathbf{B}\mathbf{A}$ with rank $\gamma$. Let $d$ denote a single data sample and $\mathcal{D}_n$ be the local dataset on device $n$. The aggregation weight $a_n=\frac{|D_n|}{|D|}$ is set as the data ratio of client $n$, where $|D_n|$ and $|D|$ represent the local and global data sizes, such that $\sum_{n=1}^N a_n=1$. The global loss function is denoted by $F(\cdot)$, and the local loss function of client $n$ is denoted by $F_n(\cdot)$. The sample-level loss function on client $n$ is denoted by $\mathcal{L}(\cdot; d)$. 

However, this conventional federated LoRA paradigm typically assumes a uniform LoRA rank $\gamma$ across all clients, which fails to account for system heterogeneity. This design can lead to significant inefficiencies: clients with limited computational and communication capabilities may become stragglers due to the high overhead of computing and transmitting LoRA updates, thereby prolonging each training round and further increasing the overall fine-tuning latency. On the other hand, clients with stronger resources may be underutilized if constrained by a small, fixed rank, resulting in insufficient model adaptation and slower convergence.

To overcome the above inefficiencies, FSLoRA develops a sketching mechanism that preserves the flexibility of LoRA while addressing system heterogeneity and resource constraints \cite{fedsketchlora}. The corresponding optimization problem can be formulated as follows:
\begin{equation}
\begin{aligned}
\min_{\mathbf{B,A}} F^S(\mathbf{B},\mathbf{A}) &= \sum_{n=1}^N a_n F_n^S(\mathbf{B},\mathbf{A}) \\
\text{where} \quad F_n^S(\mathbf{B},\mathbf{A}) &= \mathbb{E}_{d\sim \mathcal{D}_n; \mathbf{S} \sim \mathcal{S}_n} [\mathcal{L}(\mathbf{W}_0 + \mathbf{BSA};d)]. 
\end{aligned}
\end{equation}

Here, $\mathbf{S}$ denotes a sketching matrix randomly sampled from the diagonal matrix set $\mathcal{S}_n = \mathcal{S}(\gamma, k_n)$. The set $\mathcal{S}(\gamma, k_n)$ comprises diagonal matrices of size $\gamma \times \gamma$ with exactly $k_n$ non-zero entries. Formally, $\mathcal{S}(\gamma, k_n)$ is defined as:
\begin{equation}
\begin{aligned}
\mathcal{S}(\gamma,k_n)\!=\!\left \{ \!\mathbf{S}| \mathbf{S}\!=\!\frac{\gamma}{k_n}\!\sum_{j \in \Gamma_n }\!e_j e_j^T, \Gamma_n \!\subseteq\! \{1,...,\gamma\},|\Gamma_n |\!=\!k_n \!\right \},
\end{aligned}
\end{equation}
where $e_1, \dots, e_\gamma \in \mathbb{R}^\gamma$ are the standard unit basis vectors, and the set $\Gamma_n$ is a random subset of $[\gamma] = \{1, 2, \dots, \gamma\}$ sampled uniformly from all subsets of $[\gamma]$ with cardinality $k_n$. With $\mathbf{S}$ being a matrix sampled from $\mathcal{S}_n$, we have:
\begin{equation}
\begin{aligned}
\mathbf{BSA}=\frac{\gamma}{k_n}\sum_{j \in \Gamma_n} \mathbf{B}e_je_j^T \mathbf{A}.
\end{aligned}
\end{equation}

Here, $\Gamma_n$ denotes the index set indicating the non-zero positions in the sketching matrix $\mathbf{S}$. The operation $\mathbf{B}e_j$ selects the $j$-th column of matrix $\mathbf{B}$, while $e_j^T \mathbf{A}$ retrieves the $j$-th row of matrix $\mathbf{A}$. The sketching matrix $\mathbf{S}$ satisfies $\mathbb{E}_{\mathbf{S} \sim S_n}[\mathbf{S}] = \mathbf{I}_\gamma$, with $\mathbf{I}_\gamma$ denoting the identity matrix of dimension $\gamma$. This ensures that the approximated update $\mathbf{W}_0 + \mathbf{BSA}$ remains an unbiased estimator of the conventional LoRA update $\mathbf{W}_0 + \mathbf{BA}$.

Given a sketching matrix $\mathbf{S}$, the gradients of the loss function $\mathcal{L}(\mathbf{W}_0 + \mathbf{BSA}; d)$ with respect to the LoRA matrices $\mathbf{B}$ and $\mathbf{A}$ are expressed as follows:
\begin{equation}
\begin{aligned}
\nabla_{\mathbf{B}}\mathcal{L}(\mathbf{W}_0+\mathbf{BSA};d)&=\nabla\mathcal{L}(\mathbf{W}_0+\mathbf{BSA};d)\mathbf{A}^T\mathbf{S}^T \\
\nabla_{\mathbf{A}}\mathcal{L}(\mathbf{W}_0+\mathbf{BSA};d)&=\mathbf{S}^T\mathbf{B}^T\nabla\mathcal{L}(\mathbf{W}_0+\mathbf{BSA};d),
\end{aligned}
\end{equation}
where $\nabla_{\mathbf{B}}\mathcal{L}(\mathbf{W}_0+\mathbf{BSA};d)$, $\nabla_{\mathbf{A}}\mathcal{L}(\mathbf{W}_0+\mathbf{BSA};d)$ and $\nabla\mathcal{L}(\mathbf{W}_0+\mathbf{BSA};d)$ represent the gradients of $\mathcal{L}(\mathbf{W}_0+\mathbf{BSA};d)$ with respect to $\mathbf{B}$, $\mathbf{A}$ and $\mathbf{W}_0+\mathbf{BSA}$, respectively.

\subsection{Unbiased Federated LoRA with Independent Client Sampling}

In practical FL systems, it is often infeasible for every client device to participate in every training round due to factors such as limited computational resources or intermittent network connectivity. These decisions are typically independent across devices. While many existing studies rely on dependent sampling—assuming a fixed number of clients are sampled per round—such assumptions are not robust to real-world scenarios where fewer clients than expected may become available. In such cases, theoretical guarantees derived under dependent sampling no longer hold. Therefore, we adopt an independent sampling model, where each client participates in each round with an individual probability $q_n \in (0, 1]$, reflecting long-term participation frequencies.

We adopt the following adaptive aggregation rule in \cite{b15,b,b26} to ensure the aggregated global model remains unbiased (see the proof in Appendix A):
\begin{equation}
\begin{bmatrix}
 \mathbf{B}^{r+1}\\ \mathbf{A}^{r+1}
\end{bmatrix}
\!=\!
\begin{bmatrix}
 \mathbf{B}^{r}\\ \mathbf{A}^{r}
\end{bmatrix}
\!-\!\sum_{n=1}^N a_n \eta \frac{\mathbb{I}_r^n}{q_n}\sum_{h=1}^H \nabla\mathcal{L}(\mathbf{W}_0+\mathbf{B_n^{r,h}}\mathbf{S}_n^r\mathbf{A_n^{r,h}};d). \label{aggregation}
\end{equation}

Here, we denote $[\mathbf{B}^r; \mathbf{A}^r]$ as the global LoRA matrices at the $r$-th communication round, where $\mathbf{B}^r$ and $\mathbf{A}^r$ represent the shared low-rank decomposition parameters. Each sampled client performs $H$ local computation iterations per round. Let $h \in {1, \dots, H}$ denote the index of the local iteration. We use $\mathbf{B}_n^{r,h}$ and $\mathbf{A}_n^{r,h}$ to represent the local LoRA matrices of client $n$ at iteration $h$ of communication round $r$. The sketching matrix used by client $n$ in round $r$ is denoted as $\mathbf{S}_n^r$, which remains fixed for that client throughout the round. We consider a learning rate $\eta > 0$. At each communication round $r$, client $n$ is selected independently with probability $q_n$ via a Bernoulli trial: $\mathbb{I}_r^n \sim \text{Bernoulli}(q_n)$, where $0 < q_n \le 1$ denotes the participation probability of client $n$, and $\mathbf{q} = {q_1, \dots, q_N}$ represents the full set of sampling probabilities. If client $n$ participates in round $r$, then $\mathbb{I}_r^n = 1$; otherwise, $\mathbb{I}_r^n = 0$. To ensure that the aggregated global model remains an unbiased estimator of the full-sampled scenario under independent client sampling, each participating client's uploaded gradient must be scaled by the inverse of its sampling probability $q_n$ \cite{b15,b,b26}. This implies that when a device is frequently sampled, its contribution to the global model needs to be appropriately reduced, while conversely, when a device has a low sampling probability, its contribution to the global model needs to be increased.

In Algorithm 1, we summarize the proposed unbiased federated LoRA framework with independent client sampling. The detailed procedure of the fine-tuning process is described as follows:

1) \textbf{Initialization and Broadcast} (Line 2-3):
At the beginning of each communication round, the server generates sketching matrices $\{\mathbf{S}_n^r \sim \mathcal{S}_n\}_{n=1}^N$ for all clients, where $\mathcal{S}_n$ denotes the set of feasible sketching matrices for client $n$. These sketch matrices are then transmitted to the corresponding clients. Additionally, the server broadcasts the current global LoRA modules $[\mathbf{B}^r; \mathbf{A}^r]$ to all clients. The server also provides each client with a prescribed independent sampling probability $q_n$ to guide participation.

2) \textbf{Local Fine-tuning} (Line 4-14):
Each selected client performs $H$ local fine-tuning iterations using its received sketch matrix $S_n^r$. Specifically, in the $h$-th local update of the $r$-th round, client $n$ updates its LoRA parameters $\mathbf{B}_n^{r,h}$ and $\mathbf{A}_n^{r,h}$ guided by $S_n^r$, which can be formulated as:
\begin{equation}
\begin{aligned}
\begin{bmatrix}
 \mathbf{B}_n^{r,h+1}\\
\mathbf{A}_n^{r,h+1}
\end{bmatrix}
&=
\begin{bmatrix}
 \mathbf{B}_n^{r,h}\\
\mathbf{A}_n^{r,h}
\end{bmatrix} \\
&-\eta 
\begin{bmatrix}
\nabla \mathcal{L}(\mathbf{W}_0+\mathbf{B}_n^{r,h}\mathbf{S}_n^r\mathbf{A}_n^{r,h};d)(\mathbf{A}_n^{r,h})^T(\mathbf{S}_n^r)^T\\
(\mathbf{S}_n^r)^T(\mathbf{B}_n^{r,h})^T\nabla \mathcal{L}(\mathbf{W}_0+\mathbf{B}_n^{r,h}\mathbf{S}_n^r\mathbf{A}_n^{r,h};d)
\end{bmatrix}.
\end{aligned}
\end{equation}

3) \textbf{Model Update and Aggregation} (Line 15-17):
After local updates, participating clients transmit their total local gradients of $\Delta \mathbf{B}_n^r$ and $\nabla \mathbf{A}_n^r$ to the server, which can be formulated as:
\begin{equation}
\begin{bmatrix}
\Delta \mathbf{B}_n^{r,h}\\
\Delta \mathbf{A}_n^{r,h}
\end{bmatrix}
=
\begin{bmatrix}
 \mathbf{B}_n^{r,0}\\
\mathbf{A}_n^{r,0}
\end{bmatrix}
-
\begin{bmatrix}
 \mathbf{B}_n^{r,H}\\
\mathbf{A}_n^{r,H}
\end{bmatrix}.
\end{equation}
The server aggregates these updates using the unbiased estimator described in Eqn. (6), accounting for the clients' individual sampling probabilities ${q_n}$, and then updates the global LoRA matrices accordingly.

\begin{algorithm}
\caption{Unbiased Federated LoRA with Independent Client Sampling}
\textbf{Input}: Sampling probabilities $\mathbf{q} = \{q_1, \dots, q_N\}$, sketching ratios $\mathbf{k}=\{k_1, \dots, k_N\}$, initial LoRA parameters $[\mathbf{B}^0; \mathbf{A}^0]$, local iterations $H$, learning rate $\eta$, lora rank $\gamma$ \\
\textbf{Output}: Final LoRA parameters $[\mathbf{B}^R; \mathbf{A}^R]$
\begin{algorithmic}[1]
\For{$r=0$ to $R-1$}
    \State \textbf{Server:} Generate sketching matrices $\{\mathbf{S}_n^r \sim \mathcal{S}_n=S(\gamma,k_n)\}_{n=1}^{N}$
    \State Broadcast $[\mathbf{B}^r; \mathbf{A}^r]$ and $\{\mathbf{S}_n^r\}_{n=1}^{N}$ to all clients
    \For{each client $n = 1$ to $N$ in parallel}
        \State Sample $\mathbb{I}_r^n \sim \text{Bernoulli}(q_n)$
        \If{$\mathbb{I}_r^n = 1$}
            \State Initialize $[\mathbf{B}_n^{r,0}, \mathbf{A}_n^{r,0}] \gets [\mathbf{B}^r, \mathbf{A}^r]$
            \For{$h = 0$ to $H-1$}
                \State Update $[\mathbf{B}_n^{r,h+1}, \mathbf{A}_n^{r,h+1}]$ using gradient of $\mathcal{L}(\mathbf{W}_0 + \mathbf{B}_n^{r,h} \mathbf{S}_n^r \mathbf{A}_n^{r,h}; d)$
            \EndFor
            \State Compute the total local gradients $\Delta \mathbf{B}_n^r$, $\Delta \mathbf{A}_n^r$
            \State Send $\frac{1}{q_n} \Delta \mathbf{B}_n^r$, $\frac{1}{q_n} \Delta \mathbf{A}_n^r$ to the server
        \EndIf
    \EndFor
    \State \textbf{Server:} Aggregate updates:
    \State \quad $\mathbf{B}^{r+1} \gets \mathbf{B}^r - \eta \sum_{n=1}^N a_n\mathbb{I}_r^n \cdot \frac{1}{q_n} \Delta \mathbf{B}_n^r$
    \State \quad $\mathbf{A}^{r+1} \gets \mathbf{A}^r - \eta \sum_{n=1}^N a_n\mathbb{I}_r^n \cdot \frac{1}{q_n} \Delta \mathbf{A}_n^r$
\EndFor
\end{algorithmic}
\end{algorithm}

\section{Convergence Analysis}
In this section, we first introduce several assumptions used throughout the analysis. Then, based on the sketching mechanism, we derive a new convergence bound for federated LoRA under arbitrary sketching ratios and independent client sampling probabilities.

To simplify the presentation while maintaining correctness, we define the following notations: 

Let $\tilde{\mathcal{L}}(\mathbf{B}, \mathbf{A}, d; \mathbf{S}) = \mathcal{L}(\mathbf{W}_0 + \mathbf{B} \mathbf{S} \mathbf{A}; d)$ denote the loss function under a given sketching matrix $\mathbf{S}$. Define $\tilde{F}_n(\mathbf{B}, \mathbf{A}; \mathbf{S}) = \mathbb{E}_{d \sim \mathcal{D}_n}[\tilde{\mathcal{L}}(\mathbf{B}, \mathbf{A}, d; \mathbf{S})]$ as the expected loss on client $n$ given $\mathbf{S}$. Let $F_n^\mathbf{S}(\mathbf{B}, \mathbf{A}) = \mathbb{E}_{\mathbf{S} \sim \mathcal{S}_n}[\tilde{F}_n(\mathbf{B}, \mathbf{A}; \mathbf{S})]$ denote the expectation over the sketching distribution. For convenience, we denote the combined LoRA parameters as $\mathbf{X} = [\mathbf{B}; \mathbf{A}]$, and rewrite all relevant functions accordingly: $F(\mathbf{X})$, $F_n(\mathbf{X})$, $F^\mathbf{S}(\mathbf{X})$, $F_n^\mathbf{S}(\mathbf{X})$, $\tilde{F}_n(\mathbf{X}; \mathbf{S})$, and $\tilde{\mathcal{L}}(\mathbf{X}, d; \mathbf{S})$. We also define the stochastic gradient $\nabla\mathcal{L}(\mathbf{W}_0+\mathbf{B_n^{r,h}}\mathbf{S}_n^r\mathbf{A_n^{r,h}};d)$ as $g_n^{r,h}$. Lastly, $\|\cdot\|$ denotes the Frobenius norm throughout the analysis.

\vspace{-3mm}
\subsection{Assumptions}
We present a new convergence bound for Algorithm 1 with non-convex loss functions. We make the following mild assumptions that are commonly adopted in existing FL works such as \cite{fedsketchlora, NEURIPS2024_8fb96e8d, 9917343}.

\textbf{Assumption 1.} $F_n(\mathbf{X})$ is differentiable and $L$-smooth, i.e., there exists a positive constant $L$ such that $\forall \mathbf{X},\mathbf{Y}$,
\begin{center}
$\| \nabla F_n(\mathbf{X})-\nabla F_n(\mathbf{Y}) \le L \| \mathbf{X} - \mathbf{Y} \|, \forall n$.
\end{center}

\textbf{Assumption 2.} The variance of the gradient $\nabla_\mathbf{X} \tilde{F}_n(\mathbf{X};\mathbf{S})$ from the sketching matrix $\mathbf{S}\sim S_n$ can be bounded as:
\begin{center}
$\mathbb{E}_{\mathbf{S} \sim S_i}\|\nabla_\mathbf{X} \tilde{F}_n(\mathbf{X};\mathbf{S})-\nabla_\mathbf{X} F_n^S(\mathbf{X})\|^2 \le \sigma^2_s , \forall n$,
\end{center}
where $F_n^S(\mathbf{X})=\mathbb{E}_{\mathbf{S \sim S_i}}[\tilde{F}_n(\mathbf{X};\mathbf{S})]$. In addition, for a given $\mathbf{S}$, the variance of the stochastic gradient $\nabla_{\mathbf{X}} \tilde{\mathcal{L}}(\mathbf{X},d;\mathbf{S})$ due to data sampling $d \sim D_n$ can be bounded as:
\begin{center}
$\mathbb{E}_{d \sim D_n}\|\nabla_\mathbf{X} \tilde{\mathcal{L}}(\mathbf{X},d;\mathbf{S})-\nabla_\mathbf{X} \tilde{F}_n^S(\mathbf{X};\mathbf{S})\|^2 \le \sigma^2_g , \forall n$,
\end{center}

\textbf{Assumption 3.} The gradient dissimilarity between the global loss $F^S(\mathbf{X})$ and each local loss $F_n^S(\mathbf{X})$ satisfies:
\begin{center}
$\|\nabla_\mathbf{X} F_n^S(\mathbf{X})-\nabla_\mathbf{X} F^S(\mathbf{X}) \|^2 \le c_h \|\nabla_\mathbf{X} F^S(\mathbf{X})\|^2 +\sigma^2_h, \forall n$,
\end{center}
where $c_h>0$ and $F^S(\mathbf{X})=\sum_{n=1}^N a_n F_n^S(\mathbf{X})$.

\subsection{Convergence Analysis}
Based on the above system model and assumptions, we present the following Theorem 1.

\textbf{Theorem 1.} When the stated assumptions are satisfied, and if the learning rate $\eta$ satisfies $\eta \le \min\left\{\frac{1}{\sqrt{18}KHL},\ \frac{1}{6KHLQ\sqrt{c_h+1}}\right\}$, where $K^2 = \max_n \left\{ \frac{\gamma^2}{k_n^2} \right\}$ captures the worst-case sketching compression ratio, and $Q^2 = \max \left\{ \sum_{n=1}^N \frac{a_n^2}{q_n} \right\}$ characterizes the effect of heterogeneous client sampling probabilities, then Algorithm 1 guarantees the following convergence behavior:
\begin{equation}
\begin{aligned}
\frac{1}{R}&\sum_{r=0}^{R-1} \mathbb{E}[\|\nabla_{\mathbf{X}} F^S(\mathbf{X}^r)\|^2] \le 4\cdot \frac{F^S(\mathbf{X}^0)-F^*}{\eta R H} \\
&+6\eta L_sN(\sigma_g^2+\sigma_s^2+\sigma_h^2)\sum_{n=1}^N \frac{a_n^2}{q_n} \\
&+ 36 L^2 \eta^2 H^2 N (\sigma_g^2+\sigma_s^2+\sigma_h^2) \sum_{n=1}^N \frac{a_n^2}{q_n}\cdot \frac{\gamma^2}{k_n^2},
\end{aligned}
\end{equation}
where $L_s$ is the smoothness parameter of $\nabla F^S(\mathbf{X})$, which is proportional to $L$.

\begin{IEEEproof}
We present a brief proof sketch here and the complete proof can be found in Appendix B. We first derive an upper bound for the difference between \(\mathbb{E}[F^S(\mathbf{X}^{r+1})]\) and \(\mathbb{E}[F^S(\mathbf{X}^{r})]\). This is achieved by considering the local stochastic gradient descent (SGD) updates, the aggregation rule, and the $L$-smoothness of the global loss function $F$. Subsequently, we proceed to further upper bound the difference within the first step. We leverage the independence between client sampling and the randomness in data sampling of SGD, employing tools such as Jensen’s inequality and Young’s inequality. Finally, we sum the inequality derived in the previous two steps over round $r$ from 0 to \(R-1\), take the total expectation, and rearrange terms to obtain the convergence bound.
\end{IEEEproof}

Note that our convergence bound in (9) holds under arbitrary and independent client sampling probabilities $\mathbf{q}$, i.e., $\sum_{n=1}^N q_n \in (0, N]$, and arbitrary sketching ratios $\mathbf{k} = \{k_1, \dots, k_N\}$, where $k_n$ corresponds to the sketching ratio of each device $n$. Based on the results in Theorem 1, we can derive the following two theoretical insights.

1) \textbf{Insights for sampling probabilities $\mathbf{q}$}: We show that the upper bound on the averaged expected global gradient decreases as the independent sampling probabilities increase. In other words, when all devices have participation probabilities $q_n=1$, i.e., full participation, the upper bound reaches its minimum value, and the required total number of rounds $R$ to reach a preset convergence threshold $\xi$ is minimized.  The upper bound in (9) further implies that in order to obtain an unbiased global model, all clients need to participate with a positive probability for model convergence, i.e., $q_n \neq 0$, for all $n$. This is because when $q_n \to 0$, it will take infinite number of rounds $R$ to achieve convergence.

2) \textbf{Insights for sketching ratios $\mathbf{k}$}: The upper bound on the average expected global gradient norm decreases as the sketching ratios increase. In other words, when all devices use the full set of LoRA parameters (i.e., $k_n = \gamma$ for all $n$), the bound reaches its minimum, and the number of communication rounds $R$ required to achieve a target convergence threshold $\xi$ is minimized. Moreover, the bound in (9) indicates that to ensure an unbiased global model update, each client must have a nonzero sketching ratio, i.e., $k_n > 0$.

\textbf{Remark 1-a (About $\mathbf{q}$):} Although increasing $q_n$ leads to faster convergence in terms of the number of communication rounds $R$ required to reach a target accuracy $\xi$, it does not necessarily reduce the total wall-clock convergence time. This is because a higher participation rate per round often results in lower bandwidth allocation per device, increasing communication time. Additionally, involving straggler devices—those with limited computational resources—can further prolong the round duration.

\textbf{Remark 1-b (About $\mathbf{k}$):} As for the sketching ratio $\mathbf{k}$, a smaller $k_n$ implies fewer parameters are involved in local computation, leading to lower computation and communication costs per round. Specifically, with fewer parameters to update, local devices compute faster, and only partial model updates need to be transmitted, resulting in reduced communication time. However, smaller sketching ratios generally require more rounds to achieve convergence.

\textbf{Remark 1-c (Trade-offs between $\mathbf{q}$ and $\mathbf{k}$)}: Importantly, $\mathbf{q}$ and $\mathbf{k}$ interact in nontrivial ways. For instance, a small $q_n$ with a large $k_n$ means that client $n$ participates infrequently, but updates a larger portion of LoRA parameters when it does. Conversely, a large $q_n$ with a small $k_n$ implies frequent participation with minimal updates per round. These trade-offs create a complex interplay between convergence rate and per-round latency.

These observations naturally lead us to the following key question:
\textit{How can we jointly optimize the independent client sampling probabilities $\mathbf{q}$ and sketching ratios $\mathbf{k}$ to minimize the total wall-clock time needed for the global model to reach a target convergence threshold?}

\section{Problem Formulation}
To solve the above key question, in this section, we first introduce the proposed adaptive bandwidth allocation scheme for FL with independent client sampling and sketching mechanism. Then, we present our formulated optimization problem on the independent sampling probabilities as well as the sketching ratios to minimize the wall-clock training time, subject to a convergence threshold constraint.
\subsection{Adaptive Bandwidth Allocation}
As shown in Fig. 1, we consider both heterogeneous communication and computation time. For the communication time, due to the system bandwidth limitation and wireless interference, we assume that the selected clients are allocated with orthogonal bandwidths. For the computation time, we assume it is a constant and is measurable for each client. 

Let $\tau_n$ denote the actual local computation time required by client $n$ to perform local LoRA matrix update under sketching ratio $k_n$, and let $t_n$ denote the actual communication time required to transmit the LoRA gradients with sketching ratio $k_n$ under unit bandwidth. Additionally, let $\hat{f}_n^{(r)}$ represent the bandwidth allocated to client $n$ during communication round $r$. We can show that for any independent sampling probability $\mathbf{q}$ and arbitrary sketching ratios $\mathbf{k}$, a minimum round time $\hat{T}^{(r)}(\mathbf{q},\mathbf{k})$ is obtained when the sampled clients complete their training round $r$ at the same time \cite{b15}. The minimum round-time $\hat{T}^{(r)}(\mathbf{q},\mathbf{k})$ can be expressed as:
\begin{equation}
    \hat{T}^{(r)}(\mathbf{q},\mathbf{k})=\tau_n+\frac{t_n}{\hat{f}^{(r)}_n},\forall n \in \mathcal{K}^{(r)}(\mathbf{q},\mathbf{k}),     \label{design}
\end{equation}
where $\mathcal{K}^{(r)}(\mathbf{q},\mathbf{k})$ represents the set of devices participating in round $r$ under sampling probability $\mathbf{q}$. Note that size and elements of $\mathcal{K}^{(r)}(\mathbf{q},\mathbf{k})$ vary across rounds, making the allocation of $\hat{f}_n^{(r)}$ to minimize the total wall-clock training time a highly intricate problem. For analytical convenience, we consider the expected bandwidth $f_n^{(r)}$ allocation instead and approximate $\hat{T}^{(r)}(\mathbf{q},\mathbf{k})$ with $T^{(r)}(\mathbf{q},\mathbf{k})$ under the expected bandwidth. For a given round $r$, from \eqref{design}, the expected total bandwidth $f_{tot}$ can be expressed~as:
\begin{equation}
    f_{tot}=\sum_{n=1}^N q_n f_n^{(r)}=\sum_{n=1}^N q_n \frac{t_n}{T^{(r)}(\mathbf{q},\mathbf{k})-\tau_n}.
    \label{ftot}
\end{equation}
The total training time $T_{\text{tot}}(\mathbf{q},\mathbf{k}, R)$ after $R$ rounds can therefore be expressed as:
\begin{equation}
    T_{tot}(\mathbf{q},\mathbf{k},R)=\sum_{r=1}^R T^{(r)}(\mathbf{q},\mathbf{k}).
    \label{Ttot}
\end{equation}

\subsection{Problem Formulation}
Our goal is to optimize the independent client sampling probabilities $\mathbf{q}$ and sketching ratios $\mathbf{k}$, to minimize the expected total training time $\mathbb{E}[T_{\text{tot}}(\mathbf{q},\mathbf{k}, R)]$, while ensuring that the average squared norm of global gradient $\frac{1}{R}\sum_{r=0}^{R-1}\mathbb{E}[||\nabla_\mathbf{X}F^S(\mathbf{X^r})||^2]$ is under a preset convergence threshold $\xi$ after $R$ rounds. This translates into the following optimization problem:
\begin{align}
\small
  \textbf{P1:} \quad &\min_{\mathbf{q},\mathbf{k},R} \mathbb{E}[T_{tot}(\mathbf{q},\mathbf{k},R)] \nonumber\\
  \text{s.t.} \quad & \frac{1}{R}\sum_{r=0}^{R-1}\mathbb{E}[\left \| \nabla_\mathbf{X} F^S(\mathbf{X}^r) \right \|^2] \le \xi, \quad R\in Z^+, \label{P1}\\
  &0<q_n \le 1, k_n \in \{0,1,2,...,r\},\quad \forall n \in \mathcal{N}. \nonumber
\end{align}

$\textbf{P1}$ is a joint and non-convex optimization problem and is difficult to solve. First, the global loss function $F^S(\mathbf{X})$ is generally defined and it is typically impossible to predict how $\mathbf{q}$, $\mathbf{k}$ and $R$ affect the global loss before actually training the model. Second, the analytical expression of the round time $T^{(r)}(\mathbf{q},\mathbf{k})$ is not available.

\section{Optimization of Independet Sampling and Sketching Mechanism}
In this section, we develop an efficient algorithm to solve \textbf{P1}. We first obtain an upper bound on the expected round time $\mathbb{E}[T^{(r)}(\mathbf{q},\mathbf{k})]$. Then, we formulate an approximate problem of \textbf{P1} based on the convergence upper bound in Theorem 1. Finally, we propose an algorithm to solve the approximate two-variable problem with low computational complexity. 

\subsection{Bounding the Expected Round Time $\mathbb{E}[T^{(r)}(\mathbf{q},\mathbf{k})]$}
From \eqref{Ttot}, if we can obtain the expected round time $\mathbb{E}[T^{(r)}(\mathbf{q}, \mathbf{k})]$, multiplying it by the total number of rounds $R$ will give us an approximation of the total training time, in the objective of \textbf{P1}. Without loss of generality, we assume that the $N$ clients are indexed such that their actual computation times satisfy $\tau_1 \le \tau_2 \le \cdots \le \tau_N$.

In the following theorem, we provide an upper bound for $\mathbb{E}[T^{(r)}(\mathbf{q}, \mathbf{k})]$.

\textbf{Theorem 2.} The expected round time $\mathbb{E}[T^{(r)}(\mathbf{q},\mathbf{k})]$ is upper bounded by
\begin{equation}
\mathbb{E}[T^{(r)}(\mathbf{q},\mathbf{k})] \le \frac{\sum_{n=1}^N q_n t_n}{f_{tot}}+\mathbb{E}[\max{\tau_n}],
\label{theorem2}
\end{equation}
where $\mathbb{E}[\max{\tau_n}]$ is the expected maximum computation time among the clients, given by:
\begin{equation}
\mathbb{E}[\max{\tau_n}]=q_N \tau_N + \sum_{n=1}^{N-1} \prod_{i=n+1}^{N} (1-q_i)q_n \tau_n \le \sum_{n=1}^N q_n \tau_n.
\label{maxtau}
\end{equation}
\begin{IEEEproof}
It can be shown that the probability of client $n$ being the slowest one is $\prod_{i=n+1}^{N} (1-q_i)q_n$. This is because client $n$ is the slowest implying that clients $n+1$ to $N$ did not participate in the training. Due to $\prod_{i=n+1}^{N} (1-q_i) \le 1$, the upper bound of $\mathbb{E}[\max{\tau_n}]$ can be established. Furthermore, we introduce $\max{\tau_n}$ to find the maximum value of (\ref{ftot}):
\begin{equation}
f_{tot}=\sum_{n=1}^N q_n \frac{t_n}{T^{(r)}(\mathbf{q},\mathbf{k})-\tau_n} \le \frac{\sum_{n=1}^N q_nt_n}{T^{(r)}(\mathbf{q},\mathbf{k})-\max{\tau_n}}.
\end{equation}
We move $f_{tot}$ to the right side of the equation and $T^{(r)}(\mathbf{q},\mathbf{k})$ to the left side, then take the expectation on both sides, resulting in \eqref{theorem2}.
\end{IEEEproof}
From the results in Theorem 2, we can derive an upper bound on $\mathbb{E}[T^{(r)}(\mathbf{q},\mathbf{k})]$ as follows:
\begin{equation}
\mathbb{E}[T^{(r)}(\mathbf{q},\mathbf{k})] \le \sum_{n=1}^N q_n(\frac{t_n}{f_{tot}}+\tau_n).
\label{upperbound2}
\end{equation}

\subsection{Communication and Computation Time Estimation under Sketching Mechanism}
In the above descriptions, we adopt the actual communication time $t_n$ and computation time $\tau_n$ for each client $n$. However, these values are generally not available prior to system deployment. To solve \textbf{P1} and determine the optimal sampling probabilities $\mathbf{q}$ and sketching ratios $\mathbf{k}$ that minimize the total wall-clock time for fine-tuning, we first introduce the following estimations for communication and computation time under the sketching mechanism.

\subsubsection{Communication Time Estimation}
Due to the sketching mechanism, each client $n$ only uploads the updated LoRA parameters to the server, which involves transmitting $k_n^2$ parameters instead of $\gamma^2$. Let $t_n^\gamma$ denote the time required for client $n$ to upload the full LoRA parameters (i.e., $k_n = \gamma$) under unit bandwidth. Since $t_n^\gamma$ depends on the client’s communication capability, it can be measured before the fine-tuning process. Then, the estimated communication time $t_n$ for sketching ratio $k_n$ can be expressed as:
\begin{equation}
t_n = t_n^\gamma \cdot \frac{k_n^2}{\gamma^2}.
\end{equation}

\subsubsection{Computation Time Estimation}
Similarly, client $n$ only needs to update $k_n^2$ LoRA parameters locally under sketching, instead of $\gamma^2$. Let $\tau_n^\gamma$ be the time needed for client $n$ to perform full LoRA updates ($k_n = \gamma$), which reflects its computation capability and can also be measured prior to training. Following the estimation approach in \cite{9762360}, the computation time $\tau_n$ for a sketching ratio $k_n$ is estimated as:
\begin{equation}
\tau_n = \tau_n^\gamma \cdot \frac{k_n^2}{\gamma^2}
\end{equation}

Therefore, the upper bound in \eqref{upperbound2} can be rewritten using the measurable parameters $\tau_n^\gamma$ and $t_n^\gamma$ as follows:
\begin{equation}
\mathbb{E}[T^{(r)}(\mathbf{q}, \mathbf{k})] \le \sum_{n=1}^N \frac{k_n^2}{\gamma^2} q_n \left( \frac{t_n^\gamma}{f_{\text{tot}}} + \tau_n^\gamma \right).
\label{upperbound3}
\end{equation}

\subsection{Approximate Optimization Problem for \textbf{P1}}
As a result, we can approximate the optimal solution to \textbf{P1} by minimizing an upper bound of the problem with Theorem 1 and 2. To simplify the expression, we denote some constant parameters using symbols $A$, $B$, $C$, and $D$, where $A=\frac{4(F^S(X^0)-F^*)}{\eta NH(\sigma_g^2+\sigma_s^2+\sigma_h^2)}>0$, $B=\frac{\xi}{N(\sigma_g^2+\sigma_s^2+\sigma_h^2)}>0$, $C=6\eta L_s>0$ and $D=36 L^2 \eta^2 H^2>0$, respectively.

Thus the approximated optimal solution can be expressed as follows:
\begin{equation}
\begin{aligned}
  \textbf{P2:} \min_{\mathbf{q},\mathbf{k}} &\frac{A}{B\!-\!C\!\sum_{n=1}^N\!\frac{a_n^2}{q_n}\!-\!D\!\sum_{n=1}^N\!\frac{a_n^2}{q_n}\frac{\gamma^2}{k_n^2}} \!\cdot\!\sum_{n=1}^N \frac{k_n^2}{\gamma^2}q_n(\frac{t_n}{f_{tot}}\!+\!\tau_n)  \\
  \text{s.t.} \quad &\frac{a_n^2(C+DK^2)N}{B}<q_n \le 1, \quad k_n \in \{0,...,\gamma\}, \\ &K^2=max_n\{\frac{\gamma^2}{k_n^2}\}, \quad \forall n \in \mathcal{N}. \label{P2}
\end{aligned}
\end{equation}

Here, we additionally require one condition: $\frac{a_n^2(C+DK^2)N}{B}<q_n \le 1$ to ensure that the denominator of \textbf{P2} is greater than zero (see the proof in appendix C). If we recall this condition when the data is evenly distributed among all devices, i.e., $a_n=\frac{1}{N}$, the condition simplifies to $\frac{C+DK^2}{NB}<q_n \le 1$.

\vspace{2mm}
\textbf{Remark 2}: Here, we impose a tighter lower bound constraint on the sampling probabilities $q_n$ for each device, which reveals the inherent relationship between sampling probabilities and sketching ratios as well as the relationship between sampling probabilities and data heterogeneity. \emph{1) Relationship between sampling probabilities and sketching ratios:} The selection of $\mathbf{q}$ and $\mathbf{k}$ cannot be arbitrary—otherwise, the fine-tuning process may fail to converge. When the sketching ratio $k_n$ of client $n$ is small, only a limited portion of its local LoRA matrix parameters participate in local updates. In this case, the sampling probability $q_n$ must be larger than a threshold that is inversely proportional to $k_n$, so that the client participates more frequently to ensure sufficient global model updates and guarantee convergence. Conversely, when $k_n$ is large, the lower bound on $q_n$ becomes smaller, meaning that even with a lower participation frequency, the client can still contribute effectively to the global model and convergence can still be achieved. \emph{2) Relationship between sampling probabilities and data heterogeneity:} If a device possesses a larger amount of data (larger $a_n$), the lower bound for participation probability needs to be higher.

\begin{algorithm}[t]
\caption{Approximately Optimal Independent Sampling and Sketching Mechanism for Federated LoRA with Data and System Heterogeneity}
\textbf{Input}: Number of clients $N$; measured computation time $\tau_n^\gamma$, measured communication time $t_n^\gamma$ for each client $n$; total bandwidth $f_{\text{tot}}$; aggregation weight $a_n = |D_n|/|D|$; initial global model $X^0$; loss target $F_s$; step size $\epsilon$; maximum LoRA rank $\gamma$; initial sketching ratio vector $\mathbf{k^0}$\\
\textbf{Output}: Approximate optimal sampling probabilities $\mathbf{q}^*$ and sketching ratios $\mathbf{k}^*$
\begin{algorithmic}[1]
\State \textbf{Parameter Estimation Phase:}
\State Server runs Algorithm 1 with four different pre-defined configurations: $\{\mathbf{q}_1,\mathbf{k}_1\}, \{\mathbf{q}_2,\mathbf{k}_2\}, \{\mathbf{q}_3,\mathbf{k}_3\}, \{\mathbf{q}_4,\mathbf{k}_4\}$;
\State Server records the required communication rounds $R_1, R_2, R_3, R_4$ until reaching target loss $F_s$;
\State Compute $A, B, C, D$ by solving \eqref{solveabcd};

\State \textbf{Alternating Optimization Phase:}

\Repeat
    \State \textbf{Step 1: Optimize $\mathbf{q}$ with fixed $\mathbf{k}$ (Initial $\mathbf{k}^0$ )}
    \For{$M(\epsilon) \!=\! M_{\min}, M_{\min} + \epsilon, M_{\min} + 2\epsilon, \ldots, M_{\max}$}
        \State Substitute parameters $N$, $\tau_n$, $t_n$, $f_{\text{tot}}$, $a_n$, $M(\epsilon)$, $A$, $B$, $C$, $D$, $\gamma$, $\mathbf{k}$ into \textbf{P3};
        \State Solve \textbf{P3} using CVX to obtain $\mathbf{q}^*(M(\epsilon))$;
    \EndFor

\State \textbf{Step 2: Optimize $\mathbf{k}$ with fixed $\mathbf{q}$ (Greedy heuristic)}
\State Initialize $k_n \gets \gamma$ for all $n \in \mathcal{N}$;
\Repeat
    \State \texttt{updated} $\gets$ \texttt{False};
    \For{each client $n \in \mathcal{N}$}
        \If{$k_n > 1$ and reducing $k_n$ decreases \textbf{P2}}
            \State $k_n \gets k_n - 1$;
            \State \texttt{updated} $\gets$ \texttt{True};
        \EndIf
    \EndFor
\Until{\texttt{updated} = \texttt{False}}

\Until{convergence of $\mathbf{q}$ and $\mathbf{k}$ }
\State \Return $\mathbf{q}^*, \mathbf{k}^*$
\end{algorithmic}
\end{algorithm}

\subsection{Optimization Algorithm for $\mathbf{q}$ and $\mathbf{k}$}
Solving \textbf{P2} poses two main challenges. The first lies in the presence of unknown parameters, namely $A$, $B$, $C$ and $D$, which are required to evaluate the convergence condition. The second challenge arises from the bi-variable non-convex nature of \textbf{P2}, where both the sampling probabilities $\mathbf{q}$ and the sketching ratios $\mathbf{k}$ are interdependent decision variables, making the optimization problem particularly difficult to solve.

\subsubsection{Estimate the unknown parameters}
\
\newline
\indent We first propose a method to estimate the unknown parameters $A$, $B$, $C$, and $D$. This approach leverages the convergence upper bound derived in Theorem 1 and involves running Algorithm 1 with four sets of fixed sampling probabilities $q_n$ and fixed sketching ratios $k_n$ as $\{\mathbf{q_1},\mathbf{k_1}\},\{\mathbf{q_2},\mathbf{k_2}\},\{\mathbf{q_3},\mathbf{k_3}\},\{\mathbf{q_4},\mathbf{k_4}\}$. These known parameter sets can be arbitrarily chosen within the feasible range. Within each set, all clients are configured with identical parameters to ensure consistency. The values of $\mathbf{k}$ and $\mathbf{q}$ should differ between sets.

We only run a limited number of rounds to reach a predefined estimation loss \(F_s\) instead the convergence threshold \(\xi\). Let's assume that $R_1,R_2,R_3,R_4$ rounds are required to reach the loss \(F_s\) with $\{\mathbf{q_1},\mathbf{k_1}\}$, $\{\mathbf{q_2},\mathbf{k_2}\}$, $\{\mathbf{q_3},\mathbf{k_3}\}$, $\{\mathbf{q_4},\mathbf{k_4}\}$, respectively. Additionally, denote the sum \(\sum_{n=1}^N \frac{a_n^2}{q_n}\) as $Y_1$, $Y_2$, $Y_3$, $Y_4$ under $\{\mathbf{q_1},\mathbf{k_1}\}$, $\{\mathbf{q_2},\mathbf{k_2}\}$, $\{\mathbf{q_3},\mathbf{k_3}\}$, $\{\mathbf{q_4},\mathbf{k_4}\}$, respectively. We also denote the sum $\sum_{n=1}^N \frac{a_n^2}{q_n}\frac{k_n^2}{\gamma^2}$ as $Z_1$, $Z_2$, $Z_3$, $Z_4$ under $\{\mathbf{q_1},\mathbf{k_1}\}$, $\{\mathbf{q_2},\mathbf{k_2}\}$, $\{\mathbf{q_3},\mathbf{k_3}\}$, $\{\mathbf{q_4},\mathbf{k_4}\}$, respectively. As a result, we obtain four sets of observations $\{R_1, Y_1, Z_1\}, \{R_2, Y_2, Z_2\}, \{R_3, Y_3, Z_3\}, \{R_4, Y_4, Z_4\}$. By substituting these into $R=\frac{A}{B-CY-DZ}.$, we can obtain a following system:
\begin{equation}
M \cdot x = 0,
\end{equation}
where $x = [A, B, C, D]^T$, and $M \in \mathbb{R}^{4 \times 4}$ is constructed as:
\begin{equation}
M = \begin{pmatrix}
\frac{1}{R_1} & -1 & Y_1 & Z_1 \\
\frac{1}{R_1} & -1 & Y_2 & Z_2 \\
\frac{1}{R_3} & -1 & Y_3 & Z_3 \\
\frac{1}{R_4} & -1 & Y_4 & Z_4
\end{pmatrix}.
 \label{solveabcd}
\end{equation}
This can be solved to estimate the unknown parameters $A$, $B$, $C$, and $D$ using Singular Value Decomposition (SVD). Specifically, we perform SVD on $M$ as: $M = U \Sigma V^T$, where $\Sigma$ is a diagonal matrix containing the singular values. The right singular vector corresponding to the smallest singular value in $\Sigma$ provides an approximate solution $x$ in the least-squares sense.
The overall estimation process corresponds to lines 1–4 of Algorithm 2.

\vspace{2mm}
\subsubsection{Solve $\mathbf{q}$ and $\mathbf{k}$}
\
\newline
\indent Next, we propose an Alternating Minimization Algorithm to solve the bi-variable non-convex problem. Firstly, with $\mathbf{k}$ fixed, we solve for $\mathbf{q}$, e.g., setting an arbitrary initial value (such as $k_n = \gamma$). With $\mathbf{k}$ fixed, problem \textbf{P2} becomes a single-variable optimization over $\mathbf{q}$. We decompose the objective of \textbf{P2} into two multipliers: $\frac{A}{B-\sum_{n=1}^N(C+D\frac{\gamma^2}{k_n^2})\frac{a_n^2}{q_n}}$ and $\sum_{n=1}^N \frac{k_n^2}{\gamma^2}q_n(\frac{t_n}{f_{tot}}+\tau_n)$. We treat the latter term as a control variable $M$, and use a convex function to approximate the former term. Define the control variable $M$ as:
\begin{equation}
M=\sum_{n=1}^N \frac{k_n^2}{\gamma^2}q_n(\frac{t_n}{f_{tot}}+\tau_n).
\label{M}
\end{equation}
where $M_{min}=N \cdot (\frac{\min{k_n}}{\max{k_n}})^2 \min(\frac{t_n}{f_{tot}}+\tau_n) \le M \le M_{max} = N \cdot \max( \frac{t_n}{f_{tot}}+\tau_n)$, for $\forall n \in \mathcal{N}$. The first term in \textbf{P2} can be convert to a convex function using the property that the harmonic mean is no greater than the arithmetic mean, given~by:
\begin{equation}
\begin{aligned}
&\frac{A}{B-\sum_{n=1}^N(C+D\frac{\gamma^2}{k_n^2})\frac{a_n^2}{q_n}} \\
&= \frac{N \cdot A}{(B-(C+D\frac{\gamma^2}{k_1^2})\frac{a_1^2 N}{q_1})+...+(B-(C+D\frac{\gamma^2}{k_n^2})\frac{a_N^2 N}{q_N})} \\
&\le \sum_{n=1}^N\frac{A \cdot q_n}{NB q_n-a_n^2N^2(C+D\frac{\gamma^2}{k_n^2})}. \label{inequality}
\end{aligned}
\end{equation}
The equality holds if and only if $\frac{A \cdot q_i}{NB q_i-a_i^2N^2(C+D\frac{\gamma^2}{k_i^2})}=\frac{A \cdot q_j}{NB q_j-a_j^2N^2(C+D\frac{\gamma^2}{k_j^2})}, \forall i,j \in \mathcal{N}, i \neq j$. Note that when $\frac{a_n^2(C+DK^2)N}{B}<q_n \le 1, \forall n \in \mathcal{N}$, the right-hand side of (\ref{inequality}) is a convex function about $\mathbf{q}$. 

Given fixed $\mathbf{k}$, using \eqref{M} and \eqref{inequality}, we convert \textbf{P2} to the following optimization problem to solve for $\mathbf{q}$:
\begin{equation}
\begin{aligned}
  \textbf{P3:} \quad &\min_{\textbf{q}}\sum_{n=1}^N\frac{A \cdot q_n}{NB q_n-a_n^2N^2(C+D\frac{\gamma^2}{k_n^2})} M\\
  \text{s.t.} \quad & M=\sum_{n=1}^N \frac{k_n^2}{\gamma^2}q_n(\frac{t_n}{f_{tot}}+\tau_n), \label{P3}\\
  \quad &\frac{a_n^2(C+DK^2)N}{B}<q_n \le 1,  \\ &K^2=max_n\{\frac{\gamma^2}{k_n^2}\}, \quad \forall n \in \mathcal{N}.
\end{aligned}
\end{equation}

For any fixed feasible value of $M \in [M_{min},M_{max}]$, \textbf{P3} is convex in \textbf{q}. Thus, we can solve \textbf{P3} with a convex optimization tool, such as CVX, while tackling the problem by employing a linear search approach with a constant step size of $\epsilon$ over the range $[M_{min}, M_{max}]$\cite{b15}. This optimization process corresponds to lines 7-11 of the proposed Algorithm~2.

Then, with $\mathbf{q}$ fixed, we solve for $\mathbf{k}$. Since $k_n \in \{1, 2, \ldots, \gamma\}$ is an integer variable, an exhaustive search over all possible combinations would be computationally expensive, especially when $N$ is large. To improve efficiency, we adopt a greedy heuristic algorithm that iteratively adjusts each $k_n$ in a coordinate-wise manner. Starting from an initial assignment (e.g., $k_n = \gamma$), we gradually reduce each $k_n$ by one if it leads to a lower objective value, and repeat this process until no further improvement is observed. This approach provides a tractable and effective method to approximate the optimal solution of \textbf{P2}.
Therefore, $\mathbf{q}$ and $\mathbf{k}$ can be solved iteratively until convergence. This optimization process corresponds to lines 12-22 of the proposed Algorithm~2.

\section{Experiments}

In this section, we present experiments to evaluate the performance of our proposed method. All experiments are conducted on a computing cluster equipped with NVIDIA A800 GPUs, each with 80 GB of memory. The number of participating mobile devices is set to 50, and the total bandwidth in the simulated wireless communication environment is set to 100 Mbps. Our model is based on Qwen2.5-1.5B-Instruct, which we fine-tune and evaluate on the Commonsense170K dataset. This dataset includes four commonsense reasoning question-answering tasks: ARC-c, ARC-e, OBQA, and SIQA. We use a stochastic gradient descent (SGD) batch size of 4, with each client performing 10 local iterations per round. The data on each device is non-i.i.d., generated according to a Dirichlet distribution with a concentration parameter of 0.1.

\begin{figure}[t]
\centering

\begin{minipage}[t]{0.492\linewidth}
    \centering
    \includegraphics[width=\linewidth]{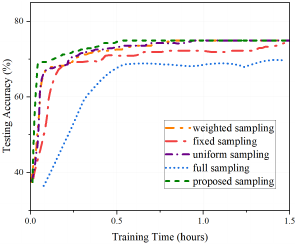}
    \caption*{(a) ARC-c}
    \label{fig:arc-c}
\end{minipage}
\hfill
\begin{minipage}[t]{0.492\linewidth}
    \centering
    \includegraphics[width=\linewidth]{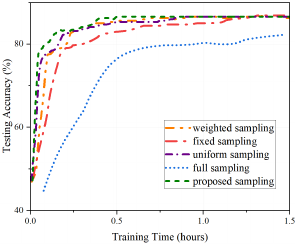}
    \caption*{(b) ARC-e}
    \label{fig:arc-e}
\end{minipage}

\begin{minipage}[t]{0.492\linewidth}
    \centering
    \includegraphics[width=\linewidth]{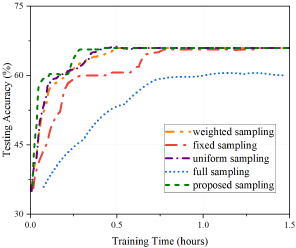}
    \caption*{(c) OBQA}
    \label{fig:obqa}
\end{minipage}
\hfill
\begin{minipage}[t]{0.492\linewidth}
    \centering
    \includegraphics[width=\linewidth]{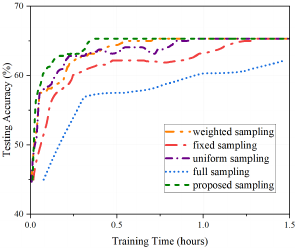}
    \caption*{(d) SIQA}
    \label{fig:siqa}
\end{minipage}

\caption{Performance analysis of the optimization of $\mathbf{q}$.}
\label{fig:benchmarks_q}
\vspace{-2mm}
\end{figure}

\subsection{Performance Analysis of the Optimization of $\mathbf{q}$}

We compare the performance of our proposed method with four benchmark client sampling strategies: (i) full sampling, where each client is selected with probability $q_n = 1$; (ii) fixed sampling, with a predefined sampling probability of $q_n = 0.2$ for all clients; (iii) uniform sampling, where $q_n = 1/N$ for all $n$; and (iv) weighted sampling, where $q_n = |D_n|/|D|$ is proportional to the size of each client’s local dataset. In all strategies, clients are sampled independently. In this set of experiments, we fix the LoRA sketching ratio for each client as $k_n = \gamma$. We perform multiple runs and report the average performance. The experimental results are presented in Fig.~\ref{fig:benchmarks_q}. Our analysis focuses on wall-clock time, with all time measurements reported in hours.

From the experimental results, we observe that our proposed sampling strategy achieves the shortest wall-clock time to reach model convergence across all tasks. In contrast, the full sampling strategy suffers from severe straggler effects, as it requires the participation of all clients in every round. This significantly prolongs the duration of each communication round, resulting in the longest overall fine-tuning latency among all methods. Fixed and uniform sampling strategies assign identical participation probabilities to all clients, without considering either data or system heterogeneity. Consequently, their training efficiency is limited. Although weighted sampling takes into account the distribution of data volumes across clients, it neglects both the feature heterogeneity of the data and the system heterogeneity, leading to suboptimal efficiency. As each client performs full LoRA updates, all the sampling methods eventually achieve similar final accuracies\footnote{Notably, the full sampling method requires significantly more time to converge and thus its final performance is not fully shown within the time range of the figure.}. However, thanks to the optimization of the sampling probabilities $\mathbf{q}$, our method reaches convergence in the shortest overall wall-clock time.

\begin{figure}[t]
\centering

\begin{minipage}[t]{0.492\linewidth}
    \centering
    \includegraphics[width=\linewidth]{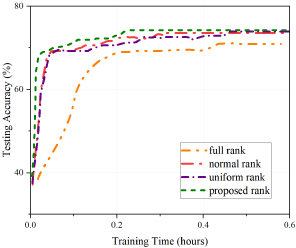}
    \caption*{(a) ARC-c}
    \label{fig:arc-c}
\end{minipage}
\hfill
\begin{minipage}[t]{0.492\linewidth}
    \centering
    \includegraphics[width=\linewidth]{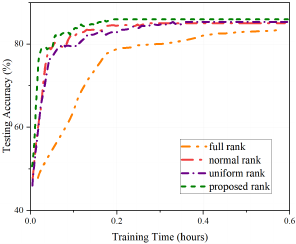}
    \caption*{(b) ARC-e}
    \label{fig:arc-e}
\end{minipage}

\begin{minipage}[t]{0.492\linewidth}
    \centering
    \includegraphics[width=\linewidth]{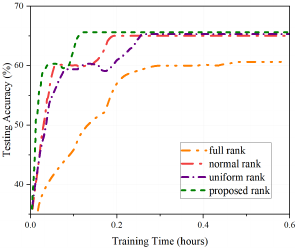}
    \caption*{(c) OBQA}
    \label{fig:obqa}
\end{minipage}
\hfill
\begin{minipage}[t]{0.492\linewidth}
    \centering
    \includegraphics[width=\linewidth]{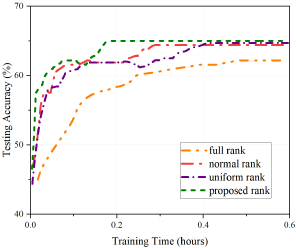}
    \caption*{(d) SIQA}
    \label{fig:siqa}
\end{minipage}

\caption{Performance analysis of the optimization of $\mathbf{k}$.}
\label{fig:benchmarks_k}
\vspace{-2mm}
\end{figure}

\subsection{Performance Analysis of the Optimization of $\mathbf{k}$}
We compare our proposed method with three different LoRA sketching ratio strategies: (i) full rank, where each client uses a sketching ratio of $k_n = \gamma$, meaning all LoRA parameters are involved in computation and communication; (ii) normal rank, where the sketching ratios $k_n \sim \mathcal{N}(\mu, \sigma^2)$ are sampled independently from a truncated normal distribution on the closed interval $[0, \gamma]$, where the mean and variance are selected to ensure values lie within the interval; (iii) uniform rank, where the sketching ratios $k_n \sim \mathcal{U}(0, \gamma)$ are sampled independently from a uniform distribution over the interval $[0, \gamma]$. In this experiment, the independent sampling probability of each client is fixed at $q_n = 0.2$. We conduct multiple runs and report the average performance. The experimental results are presented in Fig.~\ref{fig:benchmarks_k}.

From the experimental results, we observe that our proposed method consistently outperforms all other sketching ratio strategies across different datasets, demonstrating the effectiveness of our optimization approach for sketching ratios. First, our method achieves the best fine-tuning efficiency, requiring the shortest wall-clock time to reach convergence. Second, by fully considering data heterogeneity, our method attains higher final accuracy than both the normal rank and uniform rank baselines. Although the full rank strategy achieves the highest final accuracy, it suffers from significantly longer convergence time, leading to a high fine-tuning cost in practical deployments\footnote{The convergence time for full rank is too long to be displayed in the figure.}.

\begin{table}[]
\centering
\caption{Overall fine-tuning latency comparison.}
\begin{tabularx}{\linewidth}{lXXXX}
\toprule
Fine-tuning Task  & ARC-e & ARC-c & OBQA & SIQA \\
\midrule
Target Accuracy   & 80\%  & 69\%  & 60\% & 60\% \\
\midrule
\multicolumn{5}{c}{Overall Fine-tuning Time to Target Accuracy (minutes)} \\
\midrule
HeteroLoRA \cite{hetlora}          & 6.22  & 5.93  & 7.83 & 7.12 \\
FedStackLoRA \cite{flora}      & 3.60  & 3.43  & 4.55 & 4.11 \\
FSLoRA \cite{fedsketchlora}            & 3.35  & 3.20  & 4.23 & 3.83 \\
\textbf{Proposed} & \textbf{1.20} & \textbf{0.77} & \textbf{1.07} & \textbf{0.95} \\
\bottomrule
\end{tabularx}
\end{table}

\subsection{Comparison with State-Of-The-Art Methods}
In this set of experiments, we compare our proposed method with three state-of-the-art approaches: HeteroLoRA \cite{hetlora}, FedStackLoRA \cite{flora}, and FSLoRA \cite{fedsketchlora}. For each task, we define a target accuracy: 80\% for ARC-e, 69\% for ARC-c, 60\% for OBQA, and 60\% for SIQA. We evaluate the fine-tuning efficiency of each method by measuring the overall wall-clock time required to reach the corresponding target accuracy.

As shown in the results, our method consistently outperforms all baselines in terms of fine-tuning efficiency. Notably, compared to FSLoRA, our approach achieves speedups of 2.8×, 4.2×, 4.0×, and 4.0× on ARC-e, ARC-c, OBQA, and SIQA, respectively. This significant improvement is attributed to our joint optimization of independent sampling and LoRA sketching ratios, which effectively accounts for both data heterogeneity and system heterogeneity.

\section{Conclusion}
In this paper, we propose an adaptive Federated LoRA strategy with independent client sampling to minimize the convergence wall-clock time of federated fine-tuning under both computation and communication heterogeneity. We first derive a new convergence bound for Federated LoRA with arbitrary and independent client sampling. Then, we introduce an adaptive bandwidth allocation scheme that accounts for heterogeneous client resources and system bandwidth constraints. Building upon the convergence bound, we formulate and solve a non-convex optimization problem to jointly determine the LoRA sketching ratios and sampling probabilities, aiming to minimize wall-clock convergence time. We further uncover the intricate relationships between independent sampling probabilities, sketching ratios, and both system and data heterogeneity. To address the resulting bi-variable non-convex problem, we develop an efficient algorithm with low computational complexity. Finally, extensive experiments demonstrate that our approach significantly reduces wall-clock training time compared to state-of-the-art methods across various models and datasets.

\bibliographystyle{unsrt}
\bibliography{ref}

\vspace{20mm}

\appendices
\section{Proof of unbiased aggregation with independent sampling}
In this section, we prove that the aggregation method with independent sampling in \eqref{aggregation} is unbiased.
We denote the parameter matrix as $\mathbf{X} = [\mathbf{B}; \mathbf{A}]$, and define the local stochastic gradient as $g_n^{r,h} = \nabla\mathcal{L}(\mathbf{W}_0 + \mathbf{B}_n^{r,h} \mathbf{S}_n^r \mathbf{A}_n^{r,h}; d)$. Thus, given $\mathbf{X}^r$, we can get the expectation of $\mathbf{X}^{r+1}$ as follows:
{\small
\begin{equation}
\begin{aligned}
\mathbf{X}^{r+1}&=\mathbf{X}^r-\sum_{n=1}^N \eta a_n \frac{\mathbb{I}_r^n}{q_n} \sum_{h=0}^{H-1} g_n^{r,h} \\
\mathbb{E}[\mathbf{X}^{r+1}]&=\mathbf{X}^r+\sum_{n=1}^N a_n \frac{q_n}{q_n}(\mathbf{X}_n^{r,H}-\mathbf{X}^r) \\
&=\mathbf{X}^r+\sum_{n=1}^N a_n \mathbf{X}_n^{r,H} -\sum_{n=1}^N a_n \mathbf{X}^r \\
&=\sum_{n=1}^N a_n \mathbf{X}_n^{r,H}.
\end{aligned}
\end{equation}
}
The last term in the equation corresponds to aggregating the local updates uploaded by the clients to the server, weighted by $a_n$. Therefore, we can conclude that \eqref{aggregation} is unbiased.

\section{Proof of Theorem 1}
Following Appendix D.4 in \cite{fedsketchlora}, we can obtain that $F^S(\mathbf{X})$ is $L_s$-smooth, where $L_s$ is proportional to $L$.
Thus, we have:
{\small
\begin{equation}
\begin{aligned}
\mathbb{E}[F^S(\mathbf{X}^{r+1})] &\le \mathbb{E}[F^S(\mathbf{X}^r)] \\
&-\mathbb{E}[<\nabla_\mathbf{X} F^S(\mathbf{X}^r),\sum_{n=1}^N \eta a_n \frac{\mathbb{I}_r^n}{q_n}\sum_{h=0}^{H-1}g_n^{r,h}>] \\
&+\frac{L_s \eta^2}{2}\mathbb{E}[||\sum_{n=1}^N a_n \frac{\mathbb{I}_r^n}{q_n} \sum_{h=0}^{H-1} g_n^{r,h}||^2] \label{appendixB1}
\end{aligned}
\end{equation}
}

Then, we apply Jensen’s inequality to simplify the second term on the right-hand side of the inequality, resulting in:
{\small
\begin{equation}
\begin{aligned}
&-\mathbb{E}[<\nabla_\mathbf{X} F^S(\mathbf{X}^r),\sum_{n=1}^N \eta a_n \frac{\mathbb{I}_r^n}{q_n}\sum_{h=0}^{H-1}g_n^{r,h}>] \\
&=-\eta H \mathbb{E}[<\nabla_\mathbf{X} F^S(\mathbf{X}^r), \frac{N}{NH} \sum_{n=1}^N a_n \frac{\mathbb{I}_r^n}{q_n}\sum_{h=0}^{H-1} \nabla_{\mathbf{X}} F_n^S(\mathbf{X}_n^{r,h})>] \\
&=\!-\!\frac{\eta H}{2} \mathbb{E}[||\nabla_\mathbf{X} F^S\!(\mathbf{X}^r\!)||^2]\!-\!\frac{\eta}{2H} \mathbb{E}[||\!\sum_{n=1}^N\! a_n \frac{\mathbb{I}_r^n}{q_n}\!\sum_{h=0}^{H-1}\!\nabla_{\mathbf{X}} F_n^S(\mathbf{X}_n^{r,h}\!)||^2\!]\\
&+\frac{\eta H}{2} \mathbb{E}[||\nabla_\mathbf{X} F^S(\mathbf{X}^r)-\frac{N}{NH}\sum_{n=1}^N a_n \frac{\mathbb{I}_r^n}{q_n}\sum_{h=0}^{H-1}\nabla_\mathbf{X} F_n^S (\mathbf{X}_n^{r,h})||^2]\\
& = \!-\!\frac{\eta H}{2} \mathbb{E}[||\nabla_\mathbf{X} F^S\!(\mathbf{X}^r\!)||^2]\!-\!\frac{\eta}{2H} \mathbb{E}[||\!\sum_{n=1}^N\! a_n \frac{\mathbb{I}_r^n}{q_n}\!\sum_{h=0}^{H-1}\!\nabla_{\mathbf{X}} F_n^S(\mathbf{X}_n^{r,h}\!)||^2\!]\\
&\quad \quad \quad \quad + \frac{\eta N^2 H}{2} \mathbb{E}[||\frac{1}{HN}\sum_{n=1}^N \sum_{h=0}^{H-1} a_n \frac{\mathbb{E}_r^n}{q_n} \nabla_\mathbf{\mathbf{X}} F_n^S(\mathbf{X}^r) \\
&\quad \quad \quad \quad \quad \quad \quad \quad \quad \quad  -\frac{1}{NH}\sum_{n=1}^N \sum_{h=0}^{H-1} a_n \frac{\mathbb{I}_r^n}{q_n} \nabla_{\mathbf{X}} F_n^S(\mathbf{X}_n^{r,h})||^2] \\
& \le \!-\!\frac{\eta H}{2} \mathbb{E}[||\nabla_\mathbf{X} F^S\!(\mathbf{X}^r\!)||^2]\!-\!\frac{\eta}{2H} \mathbb{E}[||\!\sum_{n=1}^N\! a_n \frac{\mathbb{I}_r^n}{q_n}\!\sum_{h=0}^{H-1}\!\nabla_{\mathbf{X}} F_n^S(\mathbf{X}_n^{r,h}\!)||^2\!] \\
&+\frac{\eta N}{2}\sum_{n=1}^N \sum_{h=0}^{H-1} \frac{a_n^2}{q_n}\mathbb{E}[||\nabla_\mathbf{X} F_n^S(\mathbf{X}^r)-\nabla_\mathbf{X} F_n^S(\mathbf{X}_n^{r,h})||^2]\\
&\le \!-\!\frac{\eta H}{2} \mathbb{E}[||\nabla_\mathbf{X} F^S\!(\mathbf{X}^r\!)||^2]\!-\!\frac{\eta}{2H} \mathbb{E}[||\!\sum_{n=1}^N\! a_n \frac{\mathbb{I}_r^n}{q_n}\!\sum_{h=0}^{H-1}\!\nabla_{\mathbf{X}} F_n^S(\mathbf{X}_n^{r,h}\!)||^2\!] \\
&+\frac{\eta N}{2} \sum_{n=1}^N \sum_{h=0}^{H-1} \frac{a_n^2}{q_n} L^2 \frac{\gamma^2}{k_n^2} \mathbb{E}[||\mathbf{X}^r-\mathbf{X}_n^{r,h}||^2],
\end{aligned}
\end{equation}}
where the last inequality follows from $L_s$-smoothness $\|\nabla_\mathbf{X} F_n^S(\mathbf{X}) - \nabla_\mathbf{X} F_n^S(\mathbf{Y})\| \le L \cdot \frac{\gamma}{k_n} \|\mathbf{X} - \mathbf{Y}\|$, which has been proven in Appendix D.4 of \cite{fedsketchlora}.

Moreover,
{\small
\begin{equation}
\begin{aligned}
&\mathbb{E}[||\sum_{n=1}^N a_n \frac{\mathbb{I}_r^n}{q_n} \sum_{h=0}^{H-1} g_n^{r,h}||^2]\\
&=\mathbb{E}[||\sum_{n=1}^N a_n \frac{\mathbb{I}_r^n}{q_n}\sum_{h=0}^{H-1}g_n^{r,h}-\sum_{n=1}^N a_n \frac{\mathbb{I}_r^n}{q_n} \sum_{h=0}^{H-1}\nabla_{\mathbf{X}}\tilde{F}_n(\mathbf{X}_n^{r,h};\mathbf{S}) \\
&\!+\!\sum_{n=1}^N \!a_n\! \frac{\mathbb{I}_n^r}{q_n\!}\sum_{h=0}^{H-1} \!\nabla_\mathbf{X} \tilde{F}_n(\mathbf{X}_n^{r,h};\mathbf{S})\!-\!\sum_{n=1}^N \!a_n\! \frac{\mathbb{I}_r^n}{q_n}\!\sum_{h=0}^{H-1}\! \nabla_\mathbf{X} F_n^S (\mathbf{X}_n^{r,h})\\
&+\sum_{n=1}^N a_n \frac{\mathbb{I}_r^n}{q_n} \sum_{h=0}^{H-1} \nabla_\mathbf{X} F_n^S(\mathbf{X}_n^{r,h})||^2]\\
&\le 3\mathbb{E}[||\sum_{n=1}^N \sum_{h=0}^{H-1} a_n \frac{\mathbb{I}_r^n}{q_n}(g_n^{r,h}-\nabla_\mathbf{X} \tilde{F}_n(\mathbf{X}_n^{r,h};\mathbf{S}))||^2]\\
&+3\mathbb{E}[||\sum_{n=1}^N \sum_{h=0}^{H-1} a_n \frac{\mathbb{I}_r^n}{q_n}(\nabla_\mathbf{X} \tilde{F}_n (\mathbf{X}_n^{r,h};\mathbf{S})-\nabla_\mathbf{X} F_n^S(\mathbf{X}_n^{r,h}))||^2]\\
&+3\mathbb{E}[||\sum_{n=1}^N \sum_{h=0}^{H-1} a_n \frac{\mathbb{I}_r^n}{q_n}\nabla_\mathbf{X} F_n^S (\mathbf{X}_n^{r,h})||^2]\\
& \le \!3NH(\sigma_g^2\!+\!\sigma_s^2)\!\sum_{n=1}^N \!\frac{a_n^2}{q_n} \!+\!3\mathbb{E}[||\!\sum_{n=1}^N \!\sum_{h=0}^{H-1} \!a_n \!\frac{\mathbb{I}_r^n}{q_n}\!\nabla_\mathbf{X} F_n^S (\mathbf{X}_n^{r,h})\!||^2].
\end{aligned}
\end{equation}}

When $\eta \le \frac{1}{\sqrt{18}KHL}$, substituting the above two terms into \eqref{appendixB1} yields:
{\small
\begin{equation}
\begin{aligned}
\mathbb{E}[F^S(\mathbf{X}^{r+1})] &\le \mathbb{E}[F^S(\mathbf{X}^r)]-\frac{\eta H}{2} \mathbb{E}[||\nabla_\mathbf{X} F^S(\mathbf{X}^r)||^2]\\
&+\frac{3}{2}\eta^2 L_s H N (\sigma_g^2+\sigma_s^2) \sum_{n=1}^N \frac{a_n^2}{q_n}\\
&+\frac{1}{2}\eta L^2 N \sum_{n=1}^N \frac{a_n^2}{q_n} \frac{\gamma^2}{k_n^2}\sum_{h=0}^{H-1} \mathbb{E}[||\mathbf{X}^r-\mathbf{X}_n^{r,h}||^2].\label{appendixB2}
\end{aligned}
\end{equation}}

Here, we set:
{\small
\begin{equation}
\begin{aligned}
T=& \sum_{n=1}^N \frac{a_n^2}{q_n} \frac{\gamma^2}{k_n^2}\sum_{h=0}^{H-1} \mathbb{E}[||\mathbf{X}^r-\mathbf{X}_n^{r,h}||^2]\\
=&\sum_{n=1}^N \frac{a_n^2}{q_n} \frac{\gamma^2}{k_n^2}\sum_{h=0}^{H-1} \mathbb{E}[||\eta \sum_{\mu=0}^{h-1} g_n^{r,\mu}||^2]\\
\le & \sum_{n=1}^N \frac{a_n^2}{q_n}\frac{\gamma^2}{k_n^2}\eta^2\sum_{h=0}^{H-1}(3\mathbb{E}[||\sum_{\mu=0}^{h-1} (g_n^{r,\mu}-\nabla_\mathbf{X} \tilde{F}_n(\mathbf{X}_n^{r,\mu};\mathbf{S}))||^2]\\
&+3\mathbb{E}[||\sum_{\mu=0}^{h-1} (\nabla_\mathbf{X} \tilde{F}_n(\mathbf{X}_n^{r,\mu};\mathbf{S})-\nabla_\mathbf{X} F_n^S(\mathbf{X}_n^{r,\mu}))||^2]\\
&+3\mathbb{E}[||\sum_{\mu=0}^{h-1}\nabla_\mathbf{X} F_n^S(\mathbf{X}_n^{r,\mu})||^2])\\
\le&  \sum_{n=1}^N \frac{a_n^2}{q_n}\frac{\gamma^2}{k_n^2} \eta^2 \sum_{h=0}^{H-1}(3H(\sigma_g^2+\sigma_s^2)\!+\!3\mathbb{E}[||\sum_{\mu=0}^{h-1}\nabla_\mathbf{X}F_n^S(\mathbf{X}_n^{r,\mu})||^2])\\
\le&\sum_{n=1}^N\! \frac{a_n^2}{q_n}\frac{\gamma^2}{k_n^2} \eta^2 \!\sum_{h=0}^{H-1}(3H(\sigma_g^2+\sigma_s^2)\!+\!3h\!\sum_{\mu=0}^{h-1}(3\mathbb{E}[||\nabla_\mathbf{X}\!F_n^S\!(\mathbf{X}_n^{r,\mu})\\
&-\nabla_\mathbf{X} F_n^S(\mathbf{X}^r)||^2]+3\mathbb{E}[||\nabla_\mathbf{X} F_n^S(\mathbf{X}^r)-\nabla_\mathbf{X} F^S(\mathbf{X}^r)||^2]\\
&+3\mathbb{E}[||\nabla_\mathbf{X}F^S(\mathbf{X}^r)||^2]))\\
\le & (3\eta^2H^2\sigma_g^2+3\eta^2H^2\sigma_s^2+9\eta^2H^3\sigma_h^2)\sum_{n=1}^N \frac{a_n^2}{q_n} \frac{\gamma^2}{k_n^2}\\
&+9\eta^2H^3(c_h+1)\sum_{n=1}^N\frac{a_n^2}{q_n}\frac{\gamma^2}{k_n^2}\mathbb{E}[||\nabla_\mathbf{X}F^S(\mathbf{X}^r)||^2]\\
&+9\eta^2H^2L^2K^2 T
\end{aligned}
\end{equation}}

Thus, when $\eta\le \frac{1}{\sqrt{18}KHL}$, we can solve $T$ as:
{\small
\begin{equation}
\begin{aligned}
T \le& 18\eta^2H^3(\sigma_g^2+\sigma_s^2+\sigma_h^2)\sum_{n=1}^N \frac{a_n^2}{q_n}\frac{\gamma^2}{k_n^2}\\
&+18\eta^2H^3(c_h+1)\sum_{n=1}^N \frac{a_n^2}{q_n}\frac{\gamma^2}{k_n^2}\mathbb{E}[||\nabla_\mathbf{X}F^S(\mathbf{X}^r)||^2]
\end{aligned}
\end{equation}}

Substituting $T$ into \eqref{appendixB2}, we obtain:
{\small
\begin{equation}
\begin{aligned}
&(\frac{\eta H}{2}-9L^2\eta^3H^3N(c_h+1)\sum_{n=1}^N \frac{a_n^2}{q_n}\frac{\gamma^2}{k_n^2})\mathbb{E}[||\nabla_\mathbf{X}F^S(\mathbf{X}^r)||^2] \\
&\le \!\mathbb{E}\![F^S\!(\mathbf{X}^r)\!]\!-\!\mathbb{E}\![F^S\!(\mathbf{X}^{r+1})\!]\!+\!\frac{3}{2}\!\eta^2\!L_s\!H\!N(\sigma_g^2\!+\!\sigma_s^2\!+\!\sigma_h^2)\!\sum_{n=1}^N\!\frac{a_n^2}{q_n}\\
&+9L^2\eta^3H^3N(\sigma_g^2+\sigma_s^2+\sigma_h^2)\sum_{n=1}^N \frac{a_n^2}{q_n}\frac{\gamma^2}{k_n^2}
\end{aligned}
\end{equation}}

When $\eta \le \frac{1}{\sqrt{18}KHL}$, we can have $\frac{\eta H}{2}-9L^2\eta^3H^3N(c_h+1)\sum_{n=1}^N \frac{a_n^2}{q_n}\frac{\gamma^2}{k_n^2} \ge \frac{\eta H}{4}$. Thus, we have:
{\small
\begin{equation}
\begin{aligned}
&\mathbb{E}[||\nabla_\mathbf{X}F^S(\mathbf{X}^r)||^2] \\
&\le \!\frac{4(\mathbb{E}\![F^S\!(\mathbf{X}^r)\!]\!-\!\mathbb{E}\![F^S\!(\mathbf{X}^{r+1})\!])}{\eta H}\!+\!6\!\eta\!L_s\!N(\sigma_g^2\!+\!\sigma_s^2\!+\!\sigma_h^2)\!\sum_{n=1}^N\!\frac{a_n^2}{q_n}\\
&+36L^2\eta^2H^2N(\sigma_g^2+\sigma_s^2+\sigma_h^2)\sum_{n=1}^N \frac{a_n^2}{q_n}\frac{\gamma^2}{k_n^2}
\end{aligned}
\end{equation}}

Then, we can rearrange the terms yields and telescope from $r = \{1, ..., R\}$ to get the Theorem 1.

\section{Proof of the new bound of $\mathbf{q}$}
To ensure that the denominator of Problem \textbf{P2} is strictly greater than zero, we impose the constraint $B - C Q^2 - D K^2 Q^2 > 0$. This implies that:
{\small
\begin{equation}
\begin{aligned}
Q^2 < \frac{B}{C + D K^2}
\end{aligned}
\end{equation}}

Since $Q^2 = \sum_{n=1}^N \frac{a_n^2}{q_n}$, a sufficient condition is:
{\small
\begin{equation}
\begin{aligned}
\sum_{n=1}^N \frac{a_n^2}{q_n} < \frac{1}{N} \sum_{n=1}^N \frac{B}{C + D K^2}
\end{aligned}
\end{equation}}

which leads to the conservative per-client constraint:
{\small
\begin{equation}
\begin{aligned}
\frac{a_n^2}{q_n} < \frac{B}{N(C + D K^2)}, \quad \forall n \in \mathcal{N}
\end{aligned}
\end{equation}}

Therefore, we only need to ensure that:
{\small
\begin{equation}
\begin{aligned}
q_n > \frac{a_n^2 N(C + D K^2)}{B}
\end{aligned}
\end{equation}}

This guarantees that the denominator in \textbf{P2} remains positive.

\vfill

\end{document}